\documentclass[letterpaper, 10 pt, conference]{ieeeconf}
\IEEEoverridecommandlockouts        
\usepackage{amsmath} % assumes amsmath package installed
\usepackage{graphicx}
\usepackage{multicol}
\usepackage[bookmarks=true]{hyperref}

\usepackage{multirow}
\usepackage{booktabs}

\let\labelindent\relax
\usepackage{enumitem}
\usepackage{algorithm}
\usepackage[noend]{algpseudocode}
\usepackage[export]{adjustbox}
\algnewcommand\algorithmicdeclare{\textbf{Assume:}}
\algnewcommand\Declare{\item[\algorithmicdeclare]}
\newcommand{\oursnsp}{{\textsc{LaMB}}}
\newcommand{\ours}{{\textsc{LaMB} }}
\newcommand{\vastar}{{\textsc{$\text{VA}^*$} }}

\usepackage{listings}
\usepackage{todonotes}

\title{\LARGE \bf
Visibility-Aware Navigation Among Movable Obstacles
}

\author{Jose Muguira-Iturralde*, Aidan Curtis*, Yilun Du, Leslie Pack Kaelbling, Tomás Lozano-Pérez% <-this % stops a space
\thanks{$^{*}$The first two authors contributed equally.
The authors are at CSAIL, MIT, USA: {\tt\small \{jmuguira, curtisa, yilundu, lpk, tlp\}@mit.edu}.}}
\begin{document}
\maketitle

\thispagestyle{empty}
\pagestyle{empty}

\begin{abstract}
In this paper, we examine the problem of visibility-aware robot navigation among movable obstacles (VANAMO). A variant of the well-known NAMO robotic planning problem, VANAMO puts additional visibility constraints on robot motion and object movability. This new problem formulation lifts the restrictive assumption that the map is fully visible and the object positions are fully known. We provide a formal definition of the VANAMO problem and propose the Look and Manipulate Backchaining (\oursnsp) algorithm for solving such problems. \ours has a simple vision-based API that makes it more easily transferable to real-world robot applications and scales to the large 3D environments. To evaluate \oursnsp, we construct a set of tasks that illustrate the complex interplay between visibility and object movability that can arise in mobile base manipulation problems in unknown environments. We show that \ours outperforms NAMO and visibility-aware motion planning approaches as well as simple combinations of them on complex manipulation problems with partial observability. 

% Long-Horizon multi-step mobile manipulation typically relies on full knowledge of the underlying scene, including object properties, relationships, and affordances in the form of a 3D scene graph. Constructing such a representation from real-world data is difficult. It requires a predetermined sequence of well-framed images with known camera poses and semantic labelling at multiple resolutions ranging from scene semantics to object semantics. Previous works have focused on 3D scene graph construction and fully observed long-horizon mobile manipulation as separate problems. However, these problem are fundamentally intertwined, and will need to be addressed jointly in any autonomous robotic system. In this work, we present a strategy for active and goal-directed construction of multi-resolution 3D scene graphs. We demonstrate an instantiation of this strategy both in simulation and on a real Kinova Movo robot system. Lastly, we evaluate the performance of our system on a number of interesting tasks such as navigation among movable obstacles, mobile base tabletop manipulation, and generalized fetch.
\end{abstract}

%%%%%%%%%%%%%%%%%%%%%%%%%%%%%%%%%%%%%%%%%%%%%%%%%%%%%%%%%%%%%%%%%%%%%%%%%%%%%%%%

\section{Introduction}

Navigation is an essential ability for mobile robots. 
Typical navigation systems use motion planning for obstacle avoidance during navigation. 
However, the goal is not always reachable directly and sometimes requires manipulation of the environment, such as opening doors, grasping obstructing objects, or pushing obstructing furniture. 
The problem of robot navigation that requires manipulation of the environment is termed NAMO (Navigation Among Movable Obstacles). 
A large body of work has studied NAMO problems and presented many algorithms for solving them~\cite{Kuffner2007NavigationAM, sk2004, Stilman2006PlanningAM, manip_stilman, Moghaddam2016PlanningRN, Nieuwenhuisen2006AnEF, Levihn2012HierarchicalDT, cheezitcube, scholz_2016}. 
While the NAMO problem in its most general formulation is NP-hard, several approaches can make theoretical guarantees under certain practical assumptions that hold in many real-world tasks.

\begin{figure}[t!]
\centering
    \includegraphics[width=1\linewidth]{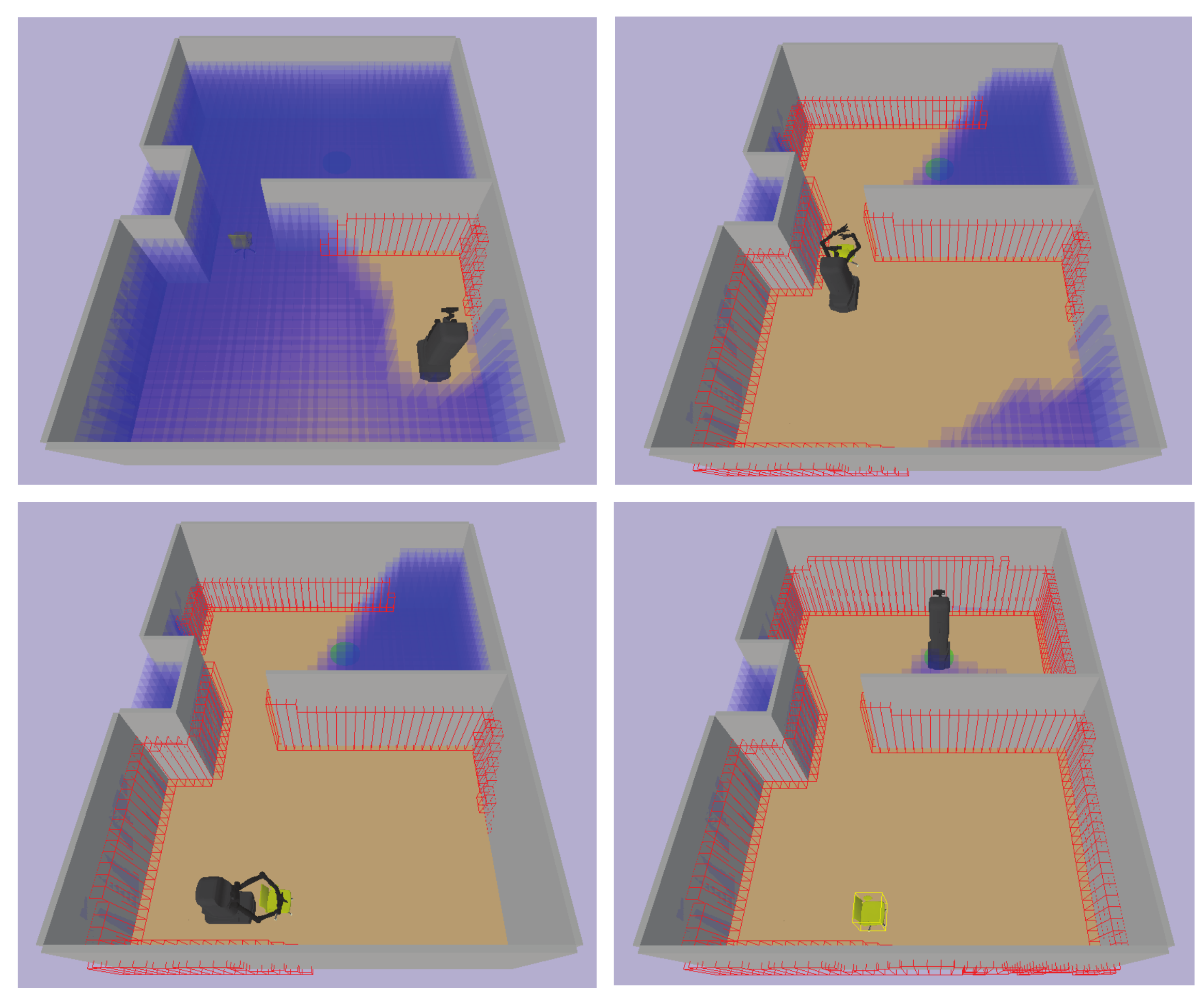}
    \caption{A partial simulated execution of a VANAMO task showing the unviewed regions (blue), observed static obstacles (red), and observed movable obstacles (yellow)}
    \label{fig:teaser}
\end{figure}

In addition to movable obstacles, another constraint that complicates navigation is visibility. For safety and reliability, a robot may not want to enter regions in the workspace that it has not observed to be free. For robots with $360^\circ$ vision and no movable obstacles, this requirement does not impose any additional constraints. However, for many robots with only a single-camera vision and an omnidirectional base (e.g. Figure~\ref{fig:teaser}), it becomes necessary to reason about where to look and where to move. Simple heuristics that force the robot to look at a region before moving into it fail when the objects being manipulated obstruct the robot's vision or when the necessary areas of the workspace are only visible from certain perspectives. Unfortunately, reasoning about visibility does not neatly fit into the motion planning problem formulation due to the path dependence of visibility. Like NAMO, visibility-aware motion planning (VAMP) is NP-hard, but several algorithms work well in practical problems~\cite{goretkin2019look, r2, r3, r4, r5}. 

Our contributions are as follows. First, we introduce a class of problems that fuses elements from NAMO and VAMP problems and formalize this class of problems as VANAMO (Visibility-Aware Navigation Among Movable Obstacles). Second, we provide a framework for mobile manipulation from camera images and implement several existing NAMO and VAMP solutions as benchmarks. Third, we propose a new algorithm for VANAMO which combines backward reasoning about visibility and reachability with forward motion planning. We demonstrate its ability to work well in complex 3D environments where the interaction between visibility and object movability plays a crucial role. Lastly, we illustrate our approach's relative effectiveness on a number of complex simulated mobile manipulation tasks.

\section{Related work}

Previous NAMO work typically relies on the concept of connected configuration regions and focuses planning effort on \textit{keyhole} actions that connect previously disconnected regions of the configuration space. Greedy backtracking algorithms have been proposed to solve the linear, or $LP_1$, class of problems in which a sequence of independent single-object keyhole actions can achieve the goal~\cite{Kuffner2007NavigationAM, sk2004, manip_stilman}. Other algorithms have extended to $LP_k$ problems that require sequencing  $k$-object keyhole actions by enforcing artificial motion constraints in the planner under additional assumptions on object movability~\cite{Stilman2006PlanningAM, Moghaddam2016PlanningRN, Nieuwenhuisen2006AnEF}. 

In many real-world robot scenarios, the environment map and object locations are not known \textit{a priori} and have to be acquired through sensors and exploration. One of the most commonly used techniques for exploration in navigation is frontier exploration~\cite{frontier}. Frontier exploration identifies a boundary between the observed and unobserved space and then picks a point on that boundary to explore next. Unfortunately, arbitrary exploration of the unknown space is inefficient when the robot has a particular navigation goal. Several methods have been proposed to address this inefficiency for visibility-aware navigation~\cite{goretkin2019look, r2, r3, r4, r5}. This problem is difficult because visibility is path-dependent and doesn't fit into the conventional motion planning framework. One recently proposed strategy develops a path-dependent motion planning algorithm that plans to view necessary regions of the workspace through subgoal backchaining~\cite{goretkin2019look}. While this algorithm applies strictly to navigation, we took inspiration from this approach in building our solution.

While the methods described so far have addressed the problems of NAMO and VAMP separately, they are insufficient for tackling the combined VANAMO problem. Robots in real-world environments must deal with visibility and movable object constraints. Several methods have been proposed to handle environments with both constraints. Some of these approaches use unrealistic models of vision that allow the agent to see through objects or assume an \textit{a priori} known map with only unknown movable object poses ~\cite{rwr, vb_stilman, snowplow}. Other approaches use realistic observations in the form of real robot sensor data but only test in simple environments where one object is obstructing the goal \cite{sensors}. To our knowledge, we are the first to present an algorithm capable of handling environments with complex constraints arising from the interaction between visibility and movable obstacles.

\begin{figure}[t!]
\centering
    \includegraphics[width=1.0\linewidth]{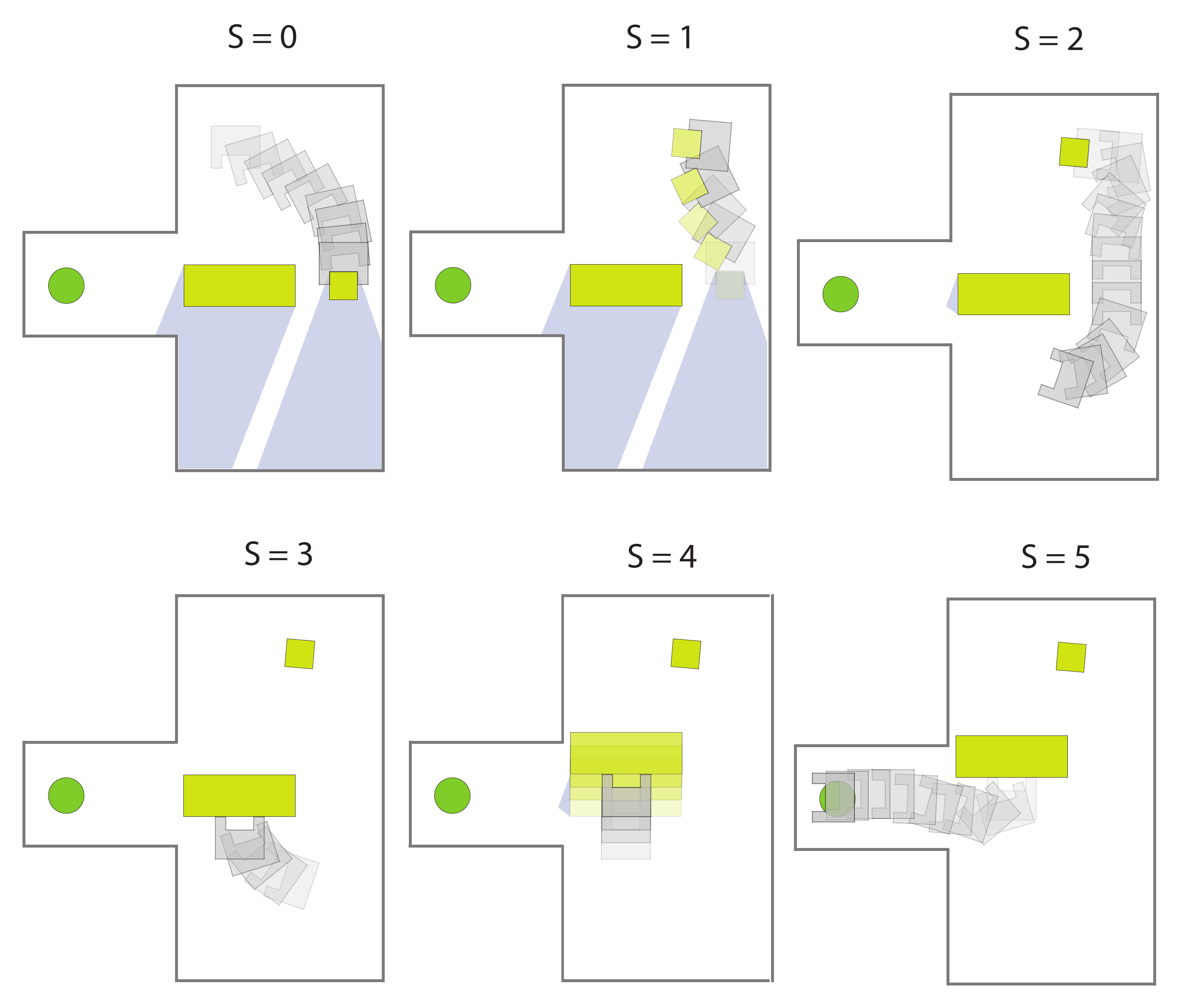}
    \caption{A top-down depiction of an example execution for the LaMB algorithm on the occluding obstacles task. The robot would like to push the wide box, but cannot because of visibility constraints. It first moves the small box out of the way (S=0, 1), then looks at the area behind the box (S=2), and then pushes the box (S=4) so it can reach the goal (S=5).}
    \label{fig:namo_complex_plan}
\end{figure}

\section{Preliminaries}
The VANAMO environment is defined by a tuple $\langle\mathcal{C}, \mathcal{W}, \mathcal{O}, \mathcal{M}, \mathcal{I}, q_g, \mathcal{A}, f, \text{Obs}\rangle$ where $\mathcal{C}$ is the robot's configuration space, $\mathcal{W}$ is the robot's workspace (typically 3D with bounds), $\mathcal{O}$ is a set of static obstacles, $\mathcal{M}$ is a set of movable objects where each $o\in\mathcal{O}$ and $m\in\mathcal{M}$ is a defined by an object shape with the same dimensionality as $\mathcal{W}$. $\mathcal{I}$ defines the world state including the initial robot configuration $q_0$ and the pose of all static obstacles $q_o$ for $o\in\mathcal{O}$ and movable obstacles $q_\mathcal{m}$ for $m\in\mathcal{M}$. The navigation goal is denoted by $q_g$. The robot also has an action space $\mathcal{A}$. The dynamics function $f:\mathcal{C}\times\mathcal{A}\rightarrow{C}$ maps from a robot configuration $q_t$ and action $a_t$ to a new configuration $q_{t+1}$. Lastly, the observation function $\text{Obs}:\mathcal{C}\rightarrow\text{Img}(q_\mathcal{M}, q_\mathcal{O})$ maps a configuration $q$ to an image projection of the environment with partial information of the movable and static objects. If the object shapes in $\mathcal{O}$ and $\mathcal{M}$ are known in advance, the observation function would be defined as $\text{Obs}:\mathcal{C}\rightarrow \mathcal{P}(\{(i, q_{i})\mid i\in\mathcal{M}\cup\mathcal{O}\})$, where $\mathcal{P}$ is the power set. 
Our experimental setting uses a holonomic robot acting in an $SE(2)$ configuration space with the degrees of freedom corresponding to $x$, $y$ movement and $\theta$ rotation. The $\text{Obs}$ function is fully defined by the intrinsics of a forward-facing camera with a fixed pose relative to the robot base.
Even with a holonomic base, explicitly modeling $\theta$ is necessary because it governs the camera direction for directional visibility. 
The robot can interact with the environment through a number of controllers. The $\texttt{move}$ controller modifies the $x$, $y$, and $\theta$ dimensions by adding or subtracting to them with fixed increments. 
The $\texttt{pick}$ controller creates a rigid attachment with a movable object if the object is within a distance $\epsilon$ of the robot, the bounding box for the movable object is smaller than a maximum height and width dimensionality (i.e., it can be surrounded by the robot arms), and some part of the object shape is within a certain $\theta$ deviation from the robot's orientation (i.e., the robot needs to be facing the object). 
The $\texttt{place}$ controller removes a rigid attachment if one exists. 
Lastly, the $\texttt{push}$ controller operates on movable objects of any size that the camera is within $\epsilon$ distance of as long as some part of the object is within a certain $\theta$ deviation from the robot's orientation. 
The $\texttt{push}$ controller displaces the robot and the movable object by a fixed distance $\Delta$ in the $q_\theta$ direction. The dynamics of the environment are assumed to be deterministic, but the algorithm may be executed in nondeterministic environments as long as replanning is triggered after major deviations from the expected outcome.

\begin{figure}[t!]
\centering
    \includegraphics[width=1\linewidth]{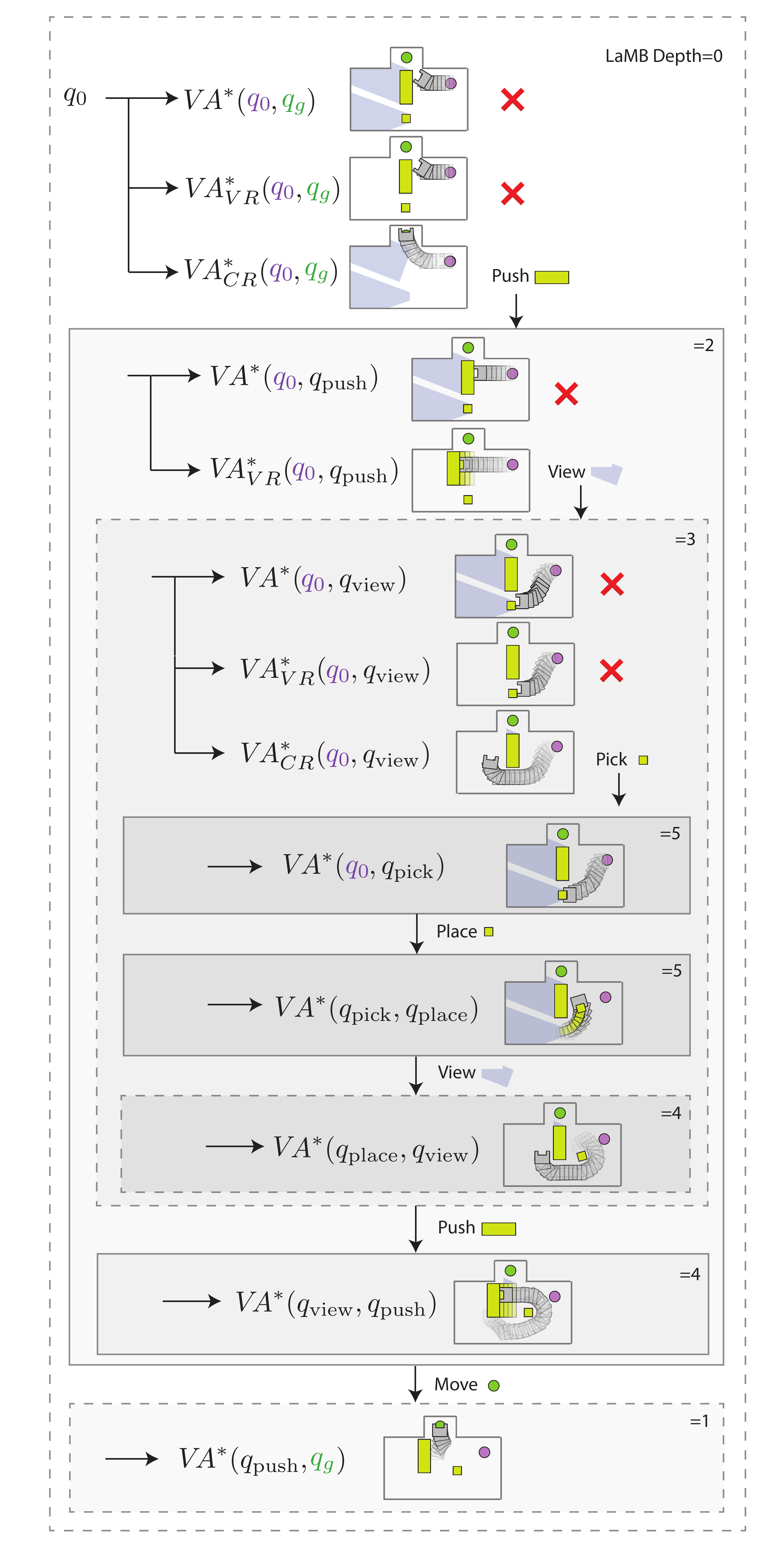}
    \caption{An example trace of the \ours algorithm on the \textbf{obstructed visibility} task depicted in Figure~\ref{fig:namo_complex_plan}. We denote each nested recursive call with a darker shade of grey. The start location is marked with a purple circle, and the goal location is marked with a green circle. $\vastar_{VR}$ denotes vision-relaxed motion planning, and $\vastar_{CR}$ denotes movable object relaxed motion planning. For clarity, we collapse successive move \& manipulate calls and denote them with solid lines.}
    \label{fig:trace}
\end{figure}

% set of parameterized controllers such as $\textsc{Move}$, $\textsc{Push}$, $\textsc{Pick}$, $\textsc{Place}$

\begin{figure*}[h]
\centering
    \includegraphics[width=1.0\linewidth]{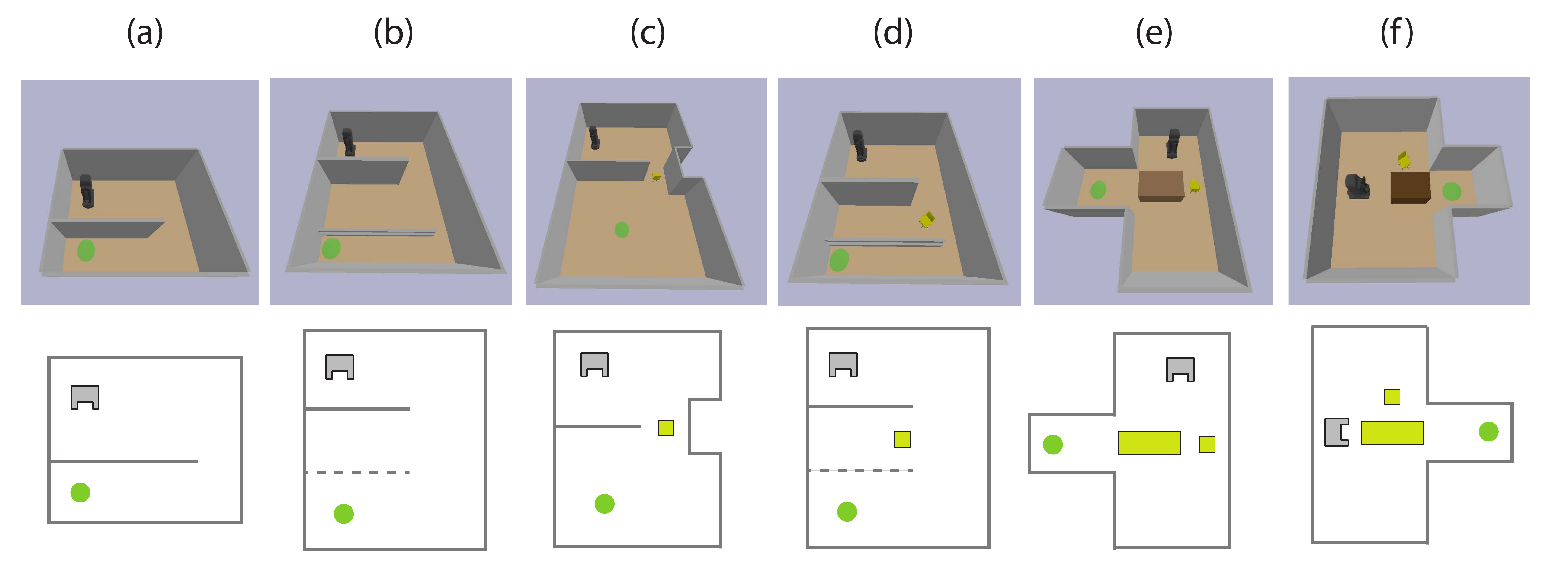}
    \caption{\textbf{Problem Instance Visualization.} A picture of problem instances from each task category (top row) with illustrated depictions of the envioronments for visual clarity (bottom row). The green dot indicates the goal position, the yellow objects in the botom row indicate movable objects. The large box is pushable, but not directly graspable.}
    \label{fig:task_categories}
\end{figure*}

\section{Method}
In this section, we introduce our algorithm for solving VANAMO problems called Look and Manipulate Backchaining (\oursnsp). Our algorithm is structured as a two-level hierarchical search. The lower level search is an A* algorithm that takes an initial configuration, goal configuration, an attachment (if one exists) and its relative pose, and a set of obstacles to avoid. The goal is not typically achievable directly through motion planning, so the higher level search finds a sequence of navigation and manipulation actions to remove the constraints preventing goal reachability. 

The constraints preventing navigation fall into two categories. \textit{Visibility constraints} ensure that the robot never passes through or moves obstacles into a region that it has not been viewed. \textit{Collision constraints} ensure that the robot never navigates or pushes obstacles through regions containing obstacles. \ours performs a high-level search through look actions (\texttt{move} actions for the purposes of looking) that remove visibility constraints and manipulation actions that displace objects and therefore change collision constraints. Because of interdependence between these configurations, it is sometimes the case that one visibility or collision constraint can prevent the robot from removing another constraint. To handle these complex interactions, our higher-level search recursively breaks down goals into subgoals that aim to remove individual constraints.
\subsection{\ours}

The \ours planner operates in the execution context outlined in algorithm~\ref{alg:lamb}. Visibility, static occupancy, and movable occupancy grids are initialized to be empty prior to any observations. 
% After a $H\times{W}$ image observation is received, the occupancy grids are updated directly from the pointcloud, which is computed from the RGB and depth images. Static occupancy and movable occupancy are separated out based on a ground truth instance segmentation provided by the simulation in the form of a $H\times{W}$ image with object body indices.
Observations have the form of segmented point clouds, and the segments are assumed to be labeled according to their object index. This segmentation includes knowledge about whether a particular object is movable. Realistically, this could be done based on material or shape properties, but in our experiments and simulations, the objects are recognized with ground truth per-pixel object identities from the simulation.
The visibility grid for a configuration, denoted $\text{Vis}(q)$, is computed by casting rays from the camera to the point specified by the captured depth image and marking all voxels each ray passes through as viewed. The \ours algorithm is then called with the updated grids as well as the navigation goal and initial state. The first action of the plan returned from \ours is executed in the environment, and the new observation is used to update the grids. This process repeats until the goal is reached or the process times out.

Inside of \ours, we consider three cases: direct planning, visibility-relaxed planning, and collision-relaxed planning, as shown in Algorithm~\ref{alg:lamb}. Each case uses $\vastar$ to determine if the goal is reachable with various relaxed constraints. $\vastar$ is a slightly modified version of $A^*$ that handles path-dependent visibility in an efficient manner. Namely, once a path to a particular state is found, we do not consider any shorter paths to that state, and we associate each state with the visibility grid that would be derived from the path that first reaches that state. This does not necessarily lead to optimal visibility-based paths but is an effective approximation~\cite{goretkin2019look}. Because the planner cannot predict what will actually be observed as the robot traverses a path, we cannot actually obtain the visibility of an imagined configuration. For this reason, we use optimistic visibility. That is, within $\vastar$, we assume that voxels not known to contain an obstacle are free space. If a direct path with $\vastar$ is possible, we simply return that path as the plan.

If no direct path is feasible, \ours first tries relaxing the visibility constraints. To do this, we simply plan with $\vastar$ and an empty visibility grid. If a plan is found with these relaxed constraints, we identify the region that needs to be viewed by intersecting the swept volume of the path, denoted $\textsc{Swept}(\text{path})$, with the already gained visibility and create a subgoal for viewing that region. It is often the case that the necessary region is not viewable from a single perspective. For this reason, we use a special type of heuristic that drives progress towards seeing part of the necessary region. We do so by computing a scalar field, F, in our workspace, which denotes the shortest distance from any point in the workspace to our required region. Given this, our new heuristic would be defined as $\text{H}_F(q) = \text{min}_{x\in \text{Vis}(q)}F(x)$. We obtain the new subgoal by running our $\vastar$ algorithm in an obstacle-relaxed environment. Each subgoal requires an independent plan, which we obtain using a recursive call to \oursnsp. This call needs to be recursive because additional constraints, such as obstructing movable obstacles, could prevent the reachability of those necessary viewing positions. In our experiments, the configuration of the robot is set so that it can always see its base. This configuration reduces the number of visibility subgoals needed to complete the task but is not strictly necessary for our algorithm to work.

If no visibility-relaxed plan is feasible, \ours computes a collision-relaxed plan that removes collision constraints imposed by \textit{movable} obstacles. To do this, we simply set the movable occupancy grid to empty and plan a path to the goal using $\vastar$. If a path is found, the first movable object collision is detected using the same $\textsc{Swept}$ subprocedure used for computing the visibility subgoal. In this work, we consider moving only the first obstacle the robot collides with along a path to the goal. This approach makes the assumption that we are dealing with $LP_1$ NAMO problems~\cite{Kuffner2007NavigationAM}. While this algorithm could easily be extended to consider multiple obstacles, doing so incurs a substantial computational cost and is not necessary for our environments. However, we do consider multiple ways of interacting with that object. For each object we consider \texttt{push} and \texttt{pick}/\texttt{place} operations (depending on the size of the object) from multiple grasp locations. Pushing is a more constrained operation but is sometimes necessary if the object is too wide for the robot to wrap its arms around. For each manipulation action considered there is a $q_{pre}$ and $q_{post}$ robot configuration. Given these intermediate configurations, we can plan a path to manipulate the object and then reach the goal through recursive calls to \ours. We additionally need to compute the updated occupancy grid and swept volume. The updated occupancy grid is necessary for planning after manipulation, and the swept volume is necessary for planning before manipulation. The swept volume adds a constraint to the planner that restricts moving other obstacles into the swept path of the manipulated object prior to the object's manipulation. See~\cite{Stilman2006PlanningAM} for details regarding this approach.

If no plan can be found under these relaxations, then we terminate the planner and return a failure result. Figure~\ref{fig:trace} shows an example trace of this algorithm on one of the more complex tasks involving multiple recursive calls with both visibility and collision relaxing.

% \begin{algorithm}
% \label{alg:lamb}
% \caption{\ours$((Q, E), W, V, q_0, Q_{goal}, v_0)$}
% \begin{algorithmic}

% \While{$q_{current} \neq Q_{goal}$}
% \State $p_{current} \gets $VAMP\_BACKCHAIN$((Q, E), W, V, q_0, Q_{goal}, v_0)$
% \While{obstruction $= False$}
%     \State $q_{c} \gets$ STEP$(q_{c} , p_{c})$
%     \State UPDATE\_VISION
%     \State obstruction $=$ FIND\_OBSTRUCTION$(q_{c} , p_{c})$

% \EndWhile
% \EndWhile

% \end{algorithmic}
% \end{algorithm}

\begin{algorithm}

\caption{$\textsc{EvaluatePlanner}(\mathcal{O}, \mathcal{M}, \mathcal{I}, q_g, \mathcal{A}, f, \text{Obs})$}
\label{alg:evaluate}
\begin{algorithmic}

\State $q_t\gets{q_0}$
\State $\text{GridV}\gets\emptyset, \text{Grid}\mathcal{O}\gets{\emptyset}, \text{Grid}\mathcal{M}\gets{\emptyset}$
\While{$\lnot({q_t}=q_g)$}
    \State $o \gets \text{Obs}(q_t)$
    \State $\text{GridV} \gets \textsc{UpdateVis}(\emptyset, o)$
    \State $\text{Grid}\mathcal{O}, \text{Grid}\mathcal{M} \gets \textsc{UpdateObs}(\text{Grid}\mathcal{O}, \text{Grid}\mathcal{M}, o)$
    \State $\text{plan}\gets\textsc{LaMB}(q_0, q_g, f, \text{Grid}\mathcal{O}, \text{Grid}\mathcal{M}, \text{GridV})$
    \State $q_t\gets{f(q_t, \text{plan}[0])}$
    \State $\text{trace}\gets{\text{trace}\oplus{(plan[0], q_t)}}$
\EndWhile
\State \Return $\text{trace}$
\end{algorithmic}
\end{algorithm}
\begin{algorithm}

\caption{$\vastar(q_0, G, \text{G}\mathcal{O}, \text{G}\mathcal{V},  \text{H}(q)=||q-q_g||_2)$}
\begin{algorithmic}

\State $Q\gets{[[q_0]]}, \text{visited}\gets{\emptyset}$
\While{$|Q|\neq{0}$}
    \State $\text{path}\gets{Q.\text{pop}(0)}, q_{\text{last}}\gets{Q[0][0]}$
    \If{$q_{\text{last}} \in G$}
        \Return \text{path}
    \EndIf
    \If{$q_{\text{last}} \in \text{visited}$}
        continue
    \EndIf
    \State $\text{V} \gets \text{PathVision}(\text{path}, \text{G}\mathcal{O})$
    \State $V^\prime \gets G\mathcal{V}\cup{V}$
    \State $N_q \gets \textsc{Neighbors}(q_{\text{last}})$
    \State $Q \gets \{ \text{path} \oplus{[q^{\prime}]}\;|\; q^\prime \in \text{N}_q, \textsc{Swept}([q'])\cap V^\prime)^c=\emptyset\}$
    \State $Q \gets \textsc{Sorted}(Q, \text{key}=\text{Cost}(\text{path}) + \text{H}(q_{\text{last}}))$
    \State$ \text{visited} \gets \text{visited}\cup\{ q_{\text{last}}\}$

\EndWhile

\end{algorithmic}
\label{alg:VA*}
\end{algorithm}

\begin{algorithm}
\caption{$\ours(q_0, q_g, \text{G}\mathcal{O}, \text{G}\mathcal{M}, \text{G}\mathcal{V})$}
\label{alg:lamb}
\begin{algorithmic}

\State $\text{plan} \gets \vastar(q_0, \{q_g\}, \text{G}\mathcal{O}\cup\text{G}\mathcal{M}, \emptyset, \text{GV})$  \Comment{Direct}
\If{${\text{plan}}$}
\State \Return $\text{plan}$
\EndIf
\State $\text{GS}\gets{\text{G}\mathcal{O}\cup\text{G}\mathcal{M}}$

\State $\text{plan} \gets \vastar(q_0, \{q_g\}, \text{G}\mathcal{O}\cup\text{G}\mathcal{M}, \emptyset)$ \Comment{Visibility-Relaxed}
\If{${\text{plan}}$}
    \State $\text{V}_{sg}\gets\textsc{Swept}(\text{plan})\cap\text{G}\mathcal{V}\,^c$
    \State $q\gets{q_0}$
    \For{$q^\prime\in \vastar(q, V_{sg}, \text{G}\mathcal{O}, \text{H}=H_F))$}
        \State $\text{plan}\gets\text{plan} \oplus\ours(q_0, q^\prime, \text{G}\mathcal{O}, \text{G}\mathcal{M}, \text{GV})$
        \State $q\gets{q^\prime}$
    \EndFor

    \State $\text{plan}\gets\text{plan} \oplus \ours(q, q_g, \text{G}\mathcal{O}, \text{G}\mathcal{M}, \text{GV})$
    \If{$\text{None}\notin\text{plan}$}
        \State \Return $\text{plan}$
    \EndIf
\EndIf

\State $\text{plan} \gets \vastar(q_0, \{q_g\}, \emptyset, \text{GV})$\Comment{Collision-Relaxed}
\If{${\text{plan}}$}
    \State $\text{Obj} \gets \textsc{FirstCollision}(\text{plan}, \text{G}\mathcal{O})$
    \State $\text{GS}\gets \text{G}\mathcal{O}\cup\text{G}\mathcal{M} \;\backslash\; \{\text{Object}\}$
    \For {$q_{pre}, q_{post} \in \textsc{SampleManip}(\text{Obj}, \text{GS})$}
        \State $\text{mid} \gets \ours(q_{pre},q_{post}, \text{G}\mathcal{O}, \text{G}\mathcal{M}\backslash{\text{Obj}}, \text{GV})$
        \State $\text{G}\mathcal{O}^\prime\gets\text{G}\mathcal{O}\cup\textsc{Swept}(\text{plan}, \text{Obj})$
        \State $\text{pre} \gets \ours(q_0, q_{pre}, \text{G}\mathcal{O}^\prime , \text{G}\mathcal{M}\backslash{\text{Obj}}, \text{GV})$
        \State $\text{G}\mathcal{O}^{\prime\prime}\gets\textsc{UpdatePose}(\text{Obj}, \text{G}\mathcal{O})$
        \State $\text{post} \gets \ours(q_{post},q_g, \text{G}\mathcal{O}^{\prime\prime}, \text{G}\mathcal{M}\backslash{\text{Obj}}, \text{GV})$
        \If{$\text{None}\notin\text{pre}\oplus\text{mid}\oplus{\text{post}}$}
            \State \Return $\text{pre}\oplus\text{mid}\oplus{\text{post}}$
        \EndIf
    \EndFor
    
\EndIf
\State \Return $\text{None}$

\end{algorithmic}
\end{algorithm}

\section{Experiments}
\label{sec:experiments}

To demonstrate the importance of visibility reasoning in NAMO and evaluate our algorithm, we construct a set of 5 task categories, each with unique challenges. The initial state for each task category can be seen in Figure~\ref{fig:task_categories}. The goal of each task category is to navigate to a particular region in space highlighted in green. Within each task category, we experiment with random initialization of object positions, robot positions, and goal region locations subject to the constraints of the task category. Below we describe each of the task categories and baselines used in our evaluation.
\subsection{Task Categories}

\textbf{Simple Navigation} is the simplest task category in which no obstacles need to be moved for the robot to reach the navigation goal (Figure~\ref{fig:task_categories}a). \textbf{Visibility} tasks are inspired by problems from visibility-aware motion planning literature~\cite{goretkin2019look, r2, r3, r4, r5}. In these tasks, it is impossible to navigate to the goal directly due to visibility constraints. (Figure~\ref{fig:task_categories}b) shows an instance wherein the robot can only move down the hallway sideways, so it must view the hallway from outside of it before moving through it to avoid collision with unseen areas of the robot's workspace. \textbf{Movable Obstacles} tasks are standard NAMO problems with movable obstacles that are fully visible from the initial state. These tasks typically require no additional visibility reasoning (Figure~\ref{fig:task_categories}c). \textbf{Obstructed Visibility} tasks have visibility constraints similar to the visibility task, but they require observations from perspectives that cannot be reached without moving one or more obstacles (Figure~\ref{fig:task_categories}e). \textbf{Occluding Obstacles} tasks involve movable objects that mostly or fully obstruct the robot's vision during interaction. Solutions to these tasks often require viewing certain regions before interacting with an object. Figure~\ref{fig:task_categories}d shows an instance and Figure~\ref{fig:namo_complex_plan} shows an example plan on that instance. Lastly, \textbf{Obstructed Affordances} tasks require manipulation of obstacles from configurations unreachable without manipulating other obstacles. Figure~\ref{fig:task_categories}f shows an example instance where the box object needs to be pushed but cannot be pushed directly because it would block the goal. It also cannot immediately be pushed from the bottom because of visibility constraints at the top of the box. The robot must move the obstructing chair and then push the box from the bottom or top.

\begin{table*}[h]
\centering
\begin{tabular}{@{}ccccccccc@{}}
\toprule
&Simple Navigation&Visibility&Movable Obstacles&Obstructed Visibility&Occluding Obstacles&Obstructed Affordance \\ \midrule
VA*      & \textbf{5/5} & 0/5 & 0/5 & 0/5 & 0/5 & 0/5 \\
NAMO     & \textbf{5/5} & 0/5 & 4/5 & 0/5 & 0/5 & 0/5 \\
FO-NAMO  & \textbf{5/5} & 0/5 & \textbf{5/5} & 0/5 & 0/5 & 0/5 \\
VAMP     & \textbf{5/5} & \textbf{5/5} & 0/5 & 0/5 & 0/5 & 0/5 \\
LaMB     & \textbf{5/5} & \textbf{5/5} & \textbf{5/5} & \textbf{5/5} & \textbf{5/5} & \textbf{5/5} \\
\bottomrule
\end{tabular}
\caption{Experimental results}
\label{table:results}
\end{table*}

\subsection{Baselines}
We compared \ours to four search baselines with visibility constraints. The \textbf{VA-Star} baseline performs an \vastar search in the discretized configuration space with a distance-to-goal heuristic. An additional constraint was added to the \vastar search that limited actions to those that did not travel through unviewed regions. The A* baseline was only able to succeed on the simple navigation task where the heuristic was a useful metric. When direction navigation to the goal was impossible, VA* would default to an exhaustive search until timeout.

The \textbf{Fully Observable NAMO} baseline is a solution to fully observable NAMO problems~\cite{Moghaddam2016PlanningRN}. This baseline first finds a relaxed path to the goal through movable obstacles. It then considers transfer paths for each obstacle in reverse order of collision starting from the goal. For each movable object the planner considers, it adds artificial collision constraints for motion on the next obstacle it tries to move. This search process is performed in a depth-first manner where infeasible motion constraints are terminal search nodes. Because of the backward planning from the goal, it is impossible to enforce visibility constraints, so this baseline assumes all unviewed space is free. Our experiments show that this baseline works when obstacles obstruct all paths to the goal but fails when visibility constraints limit obstacle motion.

The \textbf{Constrained-NAMO} baseline is inspired by related work in visual NAMO~\cite{snowplow}. This algorithm enumerates through all known visible objects, tests if moving that object will result in a shorter path to the goal, and moves the object if so. Similar to the \vastar baseline, visibility constraints are placed on the low-level configuration-space search. Our results show that this baseline fails when directly moving obstacles is impossible due to visibility or obstruction.

The \textbf{VAMP} baseline is a state-of-the-art algorithm for visibility-aware motion planning~\cite{goretkin2019look}. Similar to FO-NAMO and \ours, VAMP performs back chaining from the goal by creating intermediate subgoals based on failure from relaxed goal planning. Instead of subgoals involving movable obstacle manipulation, VAMP first tries to plan directly to the goal while relaxing vision constraints. It then identifies the regions of the workspace that were traversed through but not viewed and sets viewing those regions as a subgoal. Because VAMP does not consider setting object manipulation subgoals, it only succeeds on simple navigation and visibility tasks.

Our results show that \ours is the only algorithm capable of solving the last three tasks that each require some reasoning about the interplay between visibility and manipulation.

\subsection{Results}
Each task was run with five different seeds on each algorithm. The total success rate out of those five runs is reported in Table~\ref{table:results}. As expected, the VA* baseline was only capable of solving simple navigation tasks. VA*, unlike other motion planning algorithms, is not probabilistically complete due to visibility-based path dependence. This probabilistic incompleteness leads to definite failure when the necessary visibility constraints are not resolved via the shortest path to a configuration. Like other motion planning algorithms, $\vastar$ slows down exponentially with the increasing configuration space dimensions that come with additional movable obstacles, leading to a timeout on the NAMO problems. The Fully Observable and Constrained NAMO baselines mostly succeeded at the simple navigation and movable obstacles tasks. However, the Fully Observable baseline sometimes fails on the NAMO task because it attempts to place movable objects in unseen regions or navigate through unseen regions while holding an object. These two baselines failed at all other tasks because they did not consider visibility as a potential subgoal. The VAMP baseline succeeded at all tasks where visibility was the only constraint (simple navigation and visibility constrained) but failed when placed in an environment where moving obstacles was necessary. \ours succeeded on all seeds for each task.

\subsection{Experimental Setup}
We set up our simulated experiments in PyBullet~\cite{coumans2016pybullet}. Our robot model was a Kinova dual-arm MOVO with a head-mounted Kinect camera. All algorithms used PyBullet's built-in inverse kinematics module to determine joint positions for the base and arm configurations during manipulation and navigation. The perceptual input to each planning algorithm is the RGB, Depth, and ground truth segmentation data provided by the head-mounted camera. The image returned is $512\times{512}$ with a horizontal and vertical field of view of $90^\circ$. An example of our simulated setup can be seen in Figure~\ref{fig:teaser}. The visibility and occupancy grids are maintained in simple list structures, are updated with vectorized NumPy operations, and have a fixed resolution of 0.1 meters. All experiments were run on 6 Intel Core i7-10750H CPUs with 16GB of ram for a maximum of 2 hours before timeout. Our code is made publicly available to ensure reproducibility. \footnote{https://github.com/aidan-curtis/movo\_manipulation}

\section{Conclusion}
In this paper, we present a new problem formulation, VANAMO, that describes a class of problems for navigation with movable obstacles and partial visibility. We also proposed an algorithm, \oursnsp, that solves VANAMO problems. We demonstrate \ours on a number of complex navigation tasks that involve reasoning about visibility, object movability, and the interplay between them. Our simulated results demonstrate that \ours outperforms other baselines that do not set both navigation and manipulation subgoals. While the simplicity of the API for this algorithm lends itself to real-world use, many challenges will need to be tackled before deploying this system on a real robot. Reliable robot localization, object segmentation, movability detection, manipulation dynamics prediction, and fault tolerance will all need to be considered when deploying on a robot system. We look forward to tackling these in future work, and we hope our open-source benchmarks and algorithms will prove useful to other researchers attempting to build mobile-base manipulation systems that operate in unknown environments.

\bibliographystyle{IEEEtran}
\typeout{} % hack to overcome some overleaf/latex bug
\bibliography{references}

\end{document}